\documentclass[conference]{IEEEtran}
\usepackage{cite}
\usepackage{amsmath,amssymb,amsfonts}
\usepackage{algorithmic}
\usepackage{graphicx}
\usepackage{textcomp}
\usepackage{xcolor}

\usepackage{eurosym}
\usepackage{booktabs}

\newcommand{\refequation}[1]{Equation~\eqref{#1}}
\newcommand{\reffigure}[1]{\figurename~\ref{#1}}
\newcommand{\refsection}[1]{Section~\ref{#1}}
\newcommand{\refsubsection}[1]{Subsection~\ref{#1}}
\newcommand{\reftable}[1]{Table~\ref{#1}}

\newcommand{\argmax}{\operatornamewithlimits{argmax}}  

\def\BibTeX{{\rm B\kern-.05em{\sc i\kern-.025em b}\kern-.08em
    T\kern-.1667em\lower.7ex\hbox{E}\kern-.125emX}}
    
\begin{document}

\title{Manipulating the Distributions of Experience used for Self-Play Learning in Expert Iteration
}

\author{\IEEEauthorblockN{Dennis J. N. J. Soemers, {\'E}ric Piette, Matthew Stephenson, and Cameron Browne}
\IEEEauthorblockA{\textit{Department of Data Science and Knowledge Engineering} \\
\textit{Maastricht University}\\
Maastricht, the Netherlands \\
\texttt{\{dennis.soemers,eric.piette,matthew.stephenson,cameron.browne\}@maastrichtuniversity.nl}}
}

\IEEEoverridecommandlockouts
\IEEEpubid{\begin{minipage}{\textwidth}\ \\[12pt]
978-1-7281-4533-4/20/\$31.00 \copyright 2020 IEEE
\end{minipage}}

\maketitle

\begin{abstract}
Expert Iteration (ExIt) is an effective framework for learning game-playing policies from self-play. ExIt involves training a policy to mimic the search behaviour of a tree search algorithm -— such as Monte-Carlo tree search -— and using the trained policy to guide it. The policy and the tree search can then iteratively improve each other, through experience gathered in self-play between instances of the guided tree search algorithm. This paper outlines three different approaches for manipulating the distribution of data collected from self-play, and the procedure that samples batches for learning updates from the collected data. Firstly, samples in batches are weighted based on the durations of the episodes in which they were originally experienced. Secondly, Prioritized Experience Replay is applied within the ExIt framework, to prioritise sampling experience from which we expect to obtain valuable training signals. Thirdly, a trained exploratory policy is used to diversify the trajectories experienced in self-play. This paper summarises the effects of these manipulations on training performance evaluated in fourteen different board games. We find major improvements in early training performance in some games, and minor improvements averaged over fourteen games.
\end{abstract}


\section{Introduction} \label{Sec:Introduction}

Over the past few years, many state-of-the-art results in automated learning of policies for game-playing have been obtained by training policies using experience generated from self-play \cite{Silver2017AlphaGoZero, Anthony2017ExIt, Silver2018AlphaZero, Vinyals2019AlphaStar}. In the case of board games, the strongest results to date have been obtained using \textit{Expert Iteration} (ExIt) \cite{Silver2017AlphaGoZero, Anthony2017ExIt, Silver2018AlphaZero}, which is a self-play training framework in which an expert policy and an apprentice policy iteratively improve each other. The apprentice policy typically takes the form of a parameterised policy that can be trained, such as a neural network that outputs probability distributions over actions for given states. The expert policy is typically a search algorithm, such as \textit{Monte-Carlo tree search} (MCTS) \cite{Kocsis2006UCT, Coulom2007, Browne2012}, enhanced to use the apprentice policy to bias its search behaviour. This bias allows the apprentice policy to improve the expert policy. The expert policy subsequently improves the apprentice policy by using the searching behaviour of the expert as a training target for the apprentice policy.

In ExIt, it is customary to generate training experience by running self-play games between instances of the expert policy, where the agents select moves proportionally to the visit counts of the search processes of MCTS. In contrast to greedy move selection, selecting moves proportionally to visit counts increases the diversity of experience that can be used for training. Note that in some cases agents only select moves proportionally to visit counts in the initial portions of games to increase diversity, and switch to greedy selection in the latter parts of training games \cite{Silver2017AlphaGoZero}.

There have been numerous attempts at analysing and improving the performance of ExIt-based training procedures \cite{Tian2019ELFOpenGo, Morandin2019SAI, Wu2019AcceleratingSelfPlay}. This includes, for example, modifications to the search behaviour or architecture of the function approximator used for the policy, modification of the loss function, introduction of auxiliary targets, or other changes to the training target, and game-specific improvements (often for the game of Go) \cite{Wu2019AcceleratingSelfPlay}. Modifications to the search behaviour -- such as introducing different exploration mechanisms in the root node of MCTS -- typically lead to changes in the distribution of states that we experience, but they also affect the visit-count-based training targets. However, there has been little investigation of the role played by the distribution of data (game states encountered in self-play) that we generate, or the procedure used to sample from that experience. The most notable exceptions are publications describing state-of-the-art results in various video games \cite{Jaderberg2019HumanLevel, Vinyals2019AlphaStar}, which involved extending the notion of self-play learning to use a larger, diverse menagerie \cite{Hernandez2019Generalized} of different agents to generate experience.

In the literature on \textit{reinforcement learning} (RL) in standard single-agent settings, \textit{off-policy} RL \cite{SuttonBarto2018RLBook} is a major area of research that allows for trajectories of experience to be generated by a different \textit{behaviour policy} than the \textit{target policy} that we aim to optimise or learn something about. Among other applications, this is commonly used to generate more valuable experience to learn from through directed exploration \cite{Thrun1992RoleExploration}, or to bias the probabilities with which batches of experience are sampled based on how valuable of a training signal they are estimated to provide \cite{Schaul2016}. Similar applications may turn out to be valuable in the ExIt setting as well.

We explore three different ideas related to the manipulation of either the distribution of data, or how we sample from data, for training in ExIt -- without extending the pool of agents that generate experience to a large and diverse set \cite{Jaderberg2019HumanLevel, Vinyals2019AlphaStar}. In all three cases, we use importance sampling (IS) \cite{Kahn1953MethodsMC, Rubinstein1981SimulationMC} to correct for changes in distributions. First, we use IS in a manner that downweights samples of experience generated in longer episodes during self-play, and upweights samples of experience generated in shorter episodes. Intuitively, this makes every \emph{episode} equally ``important'' for the training objective, rather than making every \emph{game state} equally important. Second, we explore the application of \textit{Prioritized Experience Replay} \cite{Schaul2016} in ExIt. Samples of experience that are estimated to provide a valuable training signal are sampled more frequently than they would under uniform sampling, and IS is used to correct for the changed sampling strategy. Third, we train a simple policy to navigate towards game states in which the apprentice policy deviates significantly from the expert policy, and mix this policy with the standard policy that samples moves proportionally to visit counts for the purpose of move selection in self-play. This changes the distribution of data that we expect to see in the experience buffer, and we investigate the use of IS to correct for this change.

An empirical evaluation using fourteen different board games reveals major effects on training performance in individual games -- in particular improvements in early stages of training. In later stages of training, there are some games where performance degrades, but the average performance over all games is still improved. 

A formalisation of the problem setting, and background information on MCTS and IS, are provided in \refsection{Sec:Background}. \refsection{Sec:ExIt} explains implementation details of ExIt. \refsection{Sec:ImportanceSampling} discusses the use of IS in ExIt. Sections \ref{Sec:WED}, \ref{Sec:PER}, and \ref{Sec:CEE} describe the three proposed extensions. The experimental setup and results are explained in \refsection{Sec:Experiments}, and discussed in \refsection{Sec:Discussion}. Finally, \refsection{Sec:Conclusion} concludes the paper.

\section{Background} \label{Sec:Background}

In this section, we formalise the standard framework of Markov decision processes and related concepts used throughout the paper. We use bold symbols -- typically lowercase ($\boldsymbol{\pi}$, $\boldsymbol{\theta}$), but sometimes uppercase ($\boldsymbol{\mathcal{M}}$) -- to denote vectors.

\subsection{Markov decision processes}

Markov decision processes (MDPs) are a standard framework for modelling problems in which an agent perceives and acts in an environment, and is awarded rewards depending on the states it reaches and/or the actions it takes. It is commonly used throughout RL literature \cite{SuttonBarto2018RLBook}.

Every MDP consists of a finite set of states $\mathcal{S}$, a finite set of actions $\mathcal{A}$, a transition function $\mathcal{P}$, and a reward function $\mathcal{R}$. At discrete time steps $t = 0, 1, \dots$, the agent observes the current state $S_t \in \mathcal{S}$, selects an action $A_t \in \mathcal{A}$, transitions into a new state $S_{t+1}$, and observes a reward $R_{t+1}$. The transition function $\mathcal{P}$ gives the probability $\mathcal{P}(s, a, s') = \text{Pr\{} S_{t + 1} = s' \mid S_t = s, A_t = a \text{\}}$ for the agent to transition into any new state $s'$ given a previous state $S_t = s$ and a selected action $A_t = a$. Similarly, the reward function $\mathcal{R}$ gives the probability $\mathcal{R}(s, a, s', r) = \text{Pr\{} R_{t + 1} = r \mid S_t = s, A_t = a, S_{t+1} = s' \text{\}}$ for any arbitrary real number $r \in \mathbb{R}$ to be observed as a reward in that time step. 

Because it simplifies notation, we assume that every episode starts in the same initial state $S_0 = s_0$, but all the theory can trivially be extended to the case where the initial state is sampled from some fixed distribution. We are primarily interested in domains with episodes of finite length, but use sums $\sum_{t = 0}^{\infty}$ over infinite numbers of time steps throughout most of the paper -- which covers infinite-duration episodes. Finite-duration episodes, of length $T$, are still covered by setting all rewards $R_{t+1 > T}$ after $T$ time steps passed to $0$.

A policy $\pi$ is a function that, given a state $s$ and an action $a$, produces a probability $0 \leq \pi(s, a) \leq 1$ for the policy to choose to execute $a$ in $s$. Note that we require policies to yield probability distributions over all actions; $\sum_{a \in \mathcal{A}} \pi(s, a) = 1 \quad \forall s \in \mathcal{S}$. We use $\boldsymbol{\pi}(s)$ to denote a vector of probabilities for all possible entries $a \in \mathcal{A}$. We assume that all policies automatically set probabilities of any illegal actions to $0$.

The \textit{value} of a state $s$ under a policy $\pi$, denoting the (discounted) cumulative rewards that we expect to obtain when sampling actions from $\pi$ after reaching $s$, is given by;
\begin{equation}
    V^{\pi}(s) \doteq \mathbb{E} \left[ \sum_{t=0}^{\infty} \gamma^t R_{t+1} \mid A_{t' \leq t} \sim \pi \right],
\end{equation}
where $0 \leq \gamma \leq 1$ denotes a discount factor. In infinitely long episodes, we require $\gamma < 1$ to guarantee that all states have finite value. In the practical implementations and experiments described in this paper, we only use finite-duration episodes and simply take $\gamma = 1$. The notation $A_{t' \leq t}$ denotes that all actions $A_{t'}$ for $t' \leq t$ are sampled from $\pi$. Note that this expectation, and various other expectations throughout the paper, formally also depend on the choice of initial state $s_0 = S_0$. This dependence is left implicit for notational brevity.

When applying this framework to multi-player, adversarial games, we generally do so from the ``perspective'' of a single player at a time, which is oblivious to the presence of other agents and simply treats them as a part of the ``environment''. This means that states in which other players are to move are skipped over, and the influence of other agents on the probabilities with which we reach states (through their policies) is merged with the environment's transition dynamics $\mathcal{P}$. 

\subsection{Monte-Carlo tree search}

Monte-Carlo tree search (MCTS) \cite{Kocsis2006UCT, Coulom2007, Browne2012} is a tree search algorithm that gradually builds up (typically in an asymmetric fashion) its search tree over multiple iterations. During every iteration, MCTS traverses the tree that has been built up so far, using a \textit{selection} strategy that balances \textit{exploitation} of parts of the search tree that appear promising so far, and \textit{exploration} of parts of the search tree that have not yet been sufficiently explored in previous iterations. The search tree is typically expanded by a single node in the area reached by this selection strategy. A fast, (semi-)random \textit{play-out} strategy is typically used to roll out all the way to a terminal game state, which then yields a (highly noisy) estimate of the value of all states traversed in the current iteration. This value is backpropagated through the tree, and may be used to inform the selection strategy in subsequent iterations. The number of iterations that traversed through any given node during the search process is referred to as the \textit{visit count} of that node.
Note that MCTS is not restricted to the MDP framework, and can account for the actions that other agents with opposing interests may take.

\subsection{Importance sampling} \label{SubSec:ImportanceSampling}

Importance sampling (IS) \cite{Kahn1953MethodsMC, Rubinstein1981SimulationMC} is a standard technique to correct for differences between two distributions when using samples from one distribution to estimate expectations from another distribution. Suppose that we collect a set of $n$ samples $\{ x_k \mid 1 \leq k \leq n \}$ from a distribution $\mu$, and wish to estimate the expected value $\mathbb{E}_{\pi} \left[ x \right]$ under a different distribution $\pi$. Let $\mu(x_k)$ denote the probability of observing $x_k$ under $\mu$, and $\pi(x_k)$ the probability of observing $x_k$ under $\pi$. Then, the \textit{importance sampling ratios} $\rho_k = \frac{\pi(x_k)}{\mu(x_k)}$ can be used to compute an estimator $\hat{x}$ for $\mathbb{E}_{\pi} \left[ x \right]$:
\begin{equation}
    \hat{x} = \frac{\sum_{k=1}^n \rho_k x_k}{n} \approx \mathbb{E}_{\pi} \left[ x \right].
\end{equation}

This estimator is unbiased, but often exhibits high variance. This becomes particularly problematic in off-policy RL applications \cite{Precup2000EligibilityTraces, Precup2001OffPolicyTD}, where sequences of multiple IS ratios -- correcting for differences between policies across sequences of multiple time steps -- are often all multiplied together. An alternative to this estimator, referred to as the \textit{weighted importance sampling} (WIS) estimator, is given by:
\begin{equation} \label{Eq:WIS}
    \hat{x} = \frac{\sum_{k=1}^n \rho_k x_k}{\sum_{k=1}^n \rho_k}.
\end{equation}

Estimators of this form are not unbiased, but have substantially lower variance and are often found to perform better in practice -- also in off-policy RL applications \cite{Precup2000EligibilityTraces, Mahmood2014WIS}.

\section{Expert Iteration} \label{Sec:ExIt}

Expert Iteration (ExIt) \cite{Anthony2017ExIt, Silver2017AlphaGoZero} is the self-play training framework for which an intuitive description was provided in \refsection{Sec:Introduction}. This section provides a few implementation details that are particularly important for the remainder of this paper.

We aim to train a parameterised policy $\pi_{\boldsymbol{\theta}}$, with parameters $\boldsymbol{\theta}$. These are often the parameters of a deep neural network \cite{Silver2017AlphaGoZero, Anthony2017ExIt, Silver2018AlphaZero, Morandin2019SAI, Tian2019ELFOpenGo, Wu2019AcceleratingSelfPlay}, but in the empirical evaluation in this paper we focus on simpler linear function approximators. This makes it computationally feasible to perform our evaluations in \textit{general game playing} settings, using a wide variety of games as test domains. The theoretical aspects of this paper are written to facilitate either form of function approximation. Let $\boldsymbol{\phi}(s, a)$ denote a feature vector for the state-action pair $(s, a)$. For every such pair, in any given game state $s$, we compute a logit $z(s, a) = \boldsymbol{\theta}^{\top} \boldsymbol{\phi}(s, a)$. The policy's probabilities $\pi_{\boldsymbol{\theta}}(s, a)$ are then given by a softmax over all the action logits:
\begin{equation}
    \pi_{\boldsymbol{\theta}}(s, a) = \frac{\exp{z(s, a)}}{\sum_{a' \in \mathcal{A}} \exp{z(s, a')}}.
\end{equation}

Experience is generated by playing games of self-play between identical MCTS agents, which use $\pi_{\boldsymbol{\theta}}$ to guide their search. We use the same selection strategy as AlphaGo Zero \cite{Silver2017AlphaGoZero}, which traverses the tree by traversing edges that correspond to actions $a_{PUCT}$ selected according to:
\begin{equation} \label{Eq:PUCT}
    a_{PUCT} = \argmax_a \hat{Q}(s, a) + C_{PUCT} \frac{\pi_{\boldsymbol{\theta}}(s, a) \sqrt{\sum_{a'}N(s, a')}}{1 + N(s, a)},
\end{equation}
where $s$ denotes the state of the current node, $\hat{Q}(s, a)$ denotes the current value estimate of executing $a$ in $s$ as estimated by the MCTS process so far, and $N(s, a)$ denotes the visit count of the edge that is traversed by executing $a$ in $s$. Contrary to most related work with ExIt, we do not use a state-value function approximator, and only backpropagate values resulting from playouts executed using $\pi_{\boldsymbol{\theta}}$. This eliminates the need for learning a strong state-value function. 

In the self-play games, agents select moves proportional to the visit counts along edges from the root node after executing an MCTS search process for a fixed number of iterations. Suppose that we built up a search tree by running MCTS from a root node with a root state $s$. Then, we can formally define a policy $\mathcal{M}_s$, that assigns probabilities $\mathcal{M}_s(s, a)$ as follows:
\begin{equation}
\label{Eq:MCTSPolicy}
\mathcal{M}_s (s, a) = \frac{N(s, a)}{\sum_{a'} N(s, a')},
\end{equation}
where $N(s, a)$ denotes the final visit counts after searching.

Experience in self-play is generated by, for every encountered state $s$, running an MCTS process rooted in $s$, and selecting an action by sampling from $\boldsymbol{\mathcal{M}}_s(s)$. A tuple containing $s$, $\boldsymbol{\mathcal{M}}_s(s)$, and any other data required for training, is stored in a limited-capacity experience buffer that discards the oldest entries first when the maximum capacity is reached.

Training is typically done by uniformly sampling batches of experience tuples with states $s$ from the buffer, and taking gradient descent steps to minimise the cross-entropy between apprentice policy $\boldsymbol{\pi}_{\boldsymbol{\theta}}(s)$ and expert policy $\boldsymbol{\mathcal{M}}_s(s)$. The loss, estimated by averaging over a batch of size $N$, is given by:
\begin{equation} \label{Eq:CrossEntropLossEst}
    \mathcal{L}_{CE} \approx \frac{1}{N} \sum_{i=1}^N \boldsymbol{\mathcal{M}}_{s_i}(s_i)^{\top} \log \boldsymbol{\pi}_{\boldsymbol{\theta}}(s_i).
\end{equation}

It is common to include an $L_2$ regularisation term \cite{Silver2017AlphaGoZero, Silver2018AlphaZero}, but this is omitted in this paper, as our use of relatively simple function approximators and significantly lower training times reduces the risk of overfitting.

\section{Importance Sampling in ExIt} \label{Sec:ImportanceSampling}

Suppose that an experience buffer is filled with tuples of experience corresponding to all states $s_i$ encountered in self-play, as described above. If the MCTS agent used to generate experience remains fixed, the weightings $d^{\mathcal{M}}(s)$ with which we expect to observe states $s$ in the buffer is then given by:
\begin{equation}
    d^{\mathcal{M}}(s) \doteq \sum_{t=0}^{\infty} \text{Pr} \{ S_t = s \mid A_{t' < t} \sim \mathcal{M}_{S_{t'}} \}.
\end{equation}

The standard approach of sampling batches to estimate the gradients for gradient descent updates uniformly from this buffer then yields an expected probability of $p(s) = \frac{d^{\mathcal{M}}(s)}{\sum_{s'} d^{\mathcal{M}}(s')}$ for a tuple containing any particular state $s$ to be sampled. Note that the assumption that the MCTS agent used to generate experience remains fixed is a simplifying assumption. In practice, the agent's behaviour is gradually modified by updating the apprentice policy $\pi_{\boldsymbol{\theta}}$, while retaining old experience generated using older versions of the policy in the experience buffer until they are discarded due to the limited capacity of the buffer.

Sampling states according to these probabilities $p(s)$ implies that, in expectation, the cross-entropy loss that we estimate using \refequation{Eq:CrossEntropLossEst} -- and therefore optimise -- is given by:
\begin{equation} \label{Eq:CrossEntropyLoss}
    \mathcal{L}_{CE} = \sum_{s \in \mathcal{S}} p(s) \left( \boldsymbol{\mathcal{M}}_s(s)^{\top} \log \boldsymbol{\pi}_{\boldsymbol{\theta}}(s) \right).
\end{equation}

\subsection{Optimising for a different data distribution} \label{SubSec:DifferentTargetWeightings}

Generating data (experience) as described above is the most common procedure, and has produced state-of-the-art results empirically \cite{Silver2017AlphaGoZero, Silver2018AlphaZero}, but it is not certain that the optimal loss function is one that weights states $s$ by $p(s)$ as in \refequation{Eq:CrossEntropyLoss}. It is possible that different weightings may perform better. If we have \textit{target probabilities} $\omega(s)$ that we expect to work better than $p(s)$ in \refequation{Eq:CrossEntropyLoss}, we may use IS ratios $\rho(s) = \frac{\omega(s)}{p(s)}$ (as described in \refsubsection{SubSec:ImportanceSampling}) to estimate appropriate gradients -- without requiring a change in how ExIt generates experience.

\subsection{Optimising with a different data distribution}
\label{SubSec:DifferentSamplingProbabilities}

Even if we expect the cross-entropy loss function in \refequation{Eq:CrossEntropyLoss}, where states $s$ are weighted by $p(s)$, to be the optimal one to optimise. It is still possible that approaches leading to experience buffers with different data distributions, or approaches that sample from it in a different (non-uniform) manner, may be expected to perform more successfully. By using $\mu(s)$ to denote the new probability for any state $s$ to be sampled due to a modified data-generating or sampling procedure, we can specify IS ratios $\rho(s) = \frac{p(s)}{\mu(s)}$ to estimate appropriate gradients for the optimisation of \refequation{Eq:CrossEntropyLoss}. This holds even if ExIt has been modified to store (or sample from) experience in a different way.

\section{Weighting According to Episode Durations} \label{Sec:WED}

One of the original publications on ExIt \cite{Anthony2017ExIt} describes only storing \textit{a single state} $s$ in the experience buffer \textit{for every full episode} experienced in self-play. The primary motivation for this was to break correlations in the data, because states that occurred in the same episode may be highly correlated. For a similar reason, the value network of AlphaGo \cite{SilverHuangEtAl16nature} was trained from data containing only one state per game of self-play. In contrast, AlphaGo Zero \cite{Silver2017AlphaGoZero} and AlphaZero \cite{Silver2018AlphaZero} were trained using buffers that contained all states observed in self-play. Presumably, the improvements in sample efficiency were found to outweigh possible detriments due to correlated data.

Aside from the observation that storing only a single state per episode breaks correlations, it also has a different effect on the data distribution; it ensures that every episode is represented ``equally'' by a single state. When storing all states, longer-duration episodes may be argued to be ``overrepresented'' due to having more states. When storing all states in experience buffers, and therefore preserving sample efficiency, we can treat the data distribution where every episode -- regardless of duration -- is equally represented as target distribution, and use IS ratios to correct for the potential issue of overrepresentation of states from long episodes.

Let $T$ denote the duration of one particular episode. If we were to only include a single state from this episode in the experience buffer, the probability for any particular state $s$ to be selected would be $\frac{1}{T}$. Recall that $d^{\mathcal{M}}(s)$ denotes the relative weightings with which we expect to observe states $s$ when storing every state per episode, leading to probabilities $p(s)$ after dividing by $\sum_{s'} d^{\mathcal{M}}(s')$ for normalisation. The relative weightings $d^{\mathcal{M}}_{single}(s)$ with which states would be observed if we only stored a single state per episode are given by $d^{\mathcal{M}}_{single}(s) = \frac{1}{\mathbb{E}\left[ T \mid s \text{ observed} \right]} d^{\mathcal{M}}(s)$, where $\mathbb{E}\left[ T \mid s \text{ observed} \right]$ denotes the expected duration of episodes in which $s$ is observed. Normalising to probabilities leads to the following target probabilities $\omega(s)$:
\begin{equation}
\begin{aligned}
    \omega(s) &= \frac{\mathbb{T}}{\mathbb{E}\left[ T \mid s \text{ observed} \right]} \frac{d^{\mathcal{M}}(s)}{\sum_{s'} d^{\mathcal{M}}(s')} \\
    &= \frac{\mathbb{T}}{\mathbb{E}\left[ T \mid s \text{ observed} \right]} p(s),
\end{aligned}
\end{equation}
where $\mathbb{T}$ denotes the expected duration of any episode in ExIt. 

As described in \refsubsection{SubSec:DifferentTargetWeightings}, this means that we can use IS ratios $\rho(s)$ given by:
\begin{equation}
\begin{aligned}
    \rho(s) = \frac{\omega(s)}{p(s)} &= \frac{\mathbb{T}}{\mathbb{E}\left[ T \mid s \text{ observed} \right]} p(s) \frac{1}{p(s)}\\
    &= \frac{\mathbb{T}}{\mathbb{E}\left[ T \mid s \text{ observed} \right]}.
\end{aligned}
\end{equation}

In practice, the empirical duration $T$ of the episode in which any particular state $s$ was observed can be stored in the experience buffer along with $s$, and used as an unbiased estimator of $\mathbb{E}\left[ T \mid s \text{ observed} \right]$. We keep track of a moving average $\hat{\mathbb{T}}$ of episode durations during self-play as an estimator for $\mathbb{T}$. Recent episodes are given a higher weight than old episodes in this moving average, because our MCTS agent is not stationary in practice due to its use of the apprentice policy (which is trained over time). More concretely, after completing the $i^{th}$ episode with a duration $T_i$, we update $\hat{\mathbb{T}}$ as follows:
\begin{equation}
\begin{aligned}
u_i &\gets 0.95 u_{i-1} + 1 \qquad (u_0 \doteq 0) \\
\hat{\mathbb{T}} &\gets \hat{\mathbb{T}} + \frac{1}{u_i} (T_i - \hat{\mathbb{T}}).
\end{aligned}
\end{equation}

\section{Prioritized Experience Replay} \label{Sec:PER}

Prioritized Experience Replay (PER)~\cite{Schaul2016} is an approach that samples batches of experience in a non-uniform manner. Elements from a larger replay buffer are sampled more frequently if they are expected to perform a valuable training signal, and less frequently if a trained model already appears to provide accurate predictions for them. It is commonly used in value-based RL approaches, where it has been found to be one of the most valuable extensions \cite{Hessel2018Rainbow} for DQN \cite{Mnih2015DQN}.

In PER, tuples of experience $i$ in a replay (or experience) buffer are assigned \textit{priority levels} $p_i$. When sampling batches from the buffer for training, tuples $i$ are sampled with probability $P(i) = \frac{p_i^{\alpha}}{\sum_k p_k^{\alpha}}$. The exponent $\alpha$ is a hyperparameter, where $\alpha = 0$ leads to uniform sampling, and $\alpha > 0$ causes tuples with higher priority levels to be sampled more frequently. Sampling according to these probabilities can be implemented efficiently using a binary tree structure \cite{Schaul2016}.

When applied to value-based RL, priority levels are typically assigned based on the absolute values of the \textit{temporal difference} errors, which may intuitively be interpreted as the magnitudes of the mistakes made by a value function approximator for given tuples of experience. For the optimisation of the cross entropy loss (\refequation{Eq:CrossEntropyLoss}) considered in this paper, we similarly assign priorities based on the differences between apprentice and expert distributions.

Let $s_i$ denote a state that occurs in our experience buffer, with an expert distribution $\boldsymbol{\mathcal{M}}_s(s)$ over all actions, and an apprentice distribution $\boldsymbol{\pi}_{\boldsymbol{\theta}}(s)$. As in the original PER implementation \cite{Schaul2016}, the priority level is simply set equal to the maximum priority level across all existing tuples of experience if $s_i$ is newly entered (i.e. if we have not yet used it for even a single update). After using $s_i$ for an update, its new priority level $p_i$ is set by summing up the absolute differences between the distributions for all actions:
\begin{equation}
    p_i = \sum_{a \in \mathcal{A}} \left| \mathcal{M}_{s_i}(s_i, a) - \pi_{\boldsymbol{\theta}}(s_i, a) \right|.
\end{equation}

We also considered using only the maximum absolute error, rather than the sum, or simply using the cross-entropy loss $\boldsymbol{\mathcal{M}}_s(s)^{\top} \log \boldsymbol{\pi}_{\boldsymbol{\theta}}(s)$ as a priority level. We decided against using the maximum absolute error, because that tends to be a (decreasing) function of the number of legal actions in a state, more so than an indication of how well a policy performs. The cross-entropy loss was not used because its absolute value may be arbitrarily large, which can lead to instability.

As in the original PER \cite{Schaul2016}, we compute IS ratios $\rho(s_i)$ for sampled states $s_i$ using:
\begin{equation}
    \rho(s_i) = \left( \frac{1}{N} \frac{1}{P(i)} \right)^{\beta},
\end{equation}
where $N$ is the total number of tuples in the experience buffer. The exponent $\beta$ is a hyperparameter, where $\beta = 0$ leads to no corrections for bias, and $\beta = 1$ fully corrects for the changes in sampling probabilities as described in \refsubsection{SubSec:DifferentSamplingProbabilities}. For improved stability, we also divide all IS ratios $\rho(s_i)$ in any batch by the maximum IS ratio across that batch \cite{Schaul2016}.

Note that the original PER publication \cite{Schaul2016} describes multiplying the IS ratios with the temporal-difference errors in $Q$-learning updates, which yields WIS estimators \cite{Mahmood2014WIS}. In the case of the cross-entropy losses considered in this paper, we multiply the IS ratios with the full cross-entropy loss. Obtaining a WIS estimator still requires explicitly constructing an estimator of the form in \refequation{Eq:WIS}.

\section{Cross-entropy Exploration} \label{Sec:CEE}

The intuition behind PER is that states $s$ for which the apprentice policy's distribution $\boldsymbol{\pi}_{\boldsymbol{\theta}}(s)$ does not yet approximate the expert policy's distribution $\boldsymbol{\mathcal{M}}_s(s)$ may be especially valuable to learn from. This intuition does not only have to apply to the stage where we sample collected experience from a buffer, but may also inform how we should collect experience in the first place. It may be beneficial for learning to actively seek out states in self-play that lead to large differences between the two policies. We refer to this idea as Cross-Entropy Exploration (CEE).

More concretely, we train an additional policy $\mu$ using REINFORCE \cite{Williams1992REINFORCE}. At every time step $t$ in an episode, $\mu$ obtains the sum of absolute differences between probabilities assigned to all actions by the expert and apprentice as a reward:
\begin{equation}
    R_{t+1} = \sum_{a \in \mathcal{A}} \left| \mathcal{M}_{S_t}(S_t, a) - \pi_{\boldsymbol{\theta}}(S_t, a) \right|.
\end{equation}

This means that $\mu$ is trained to navigate towards states that (eventually) lead to large errors for the apprentice distribution. Note that -- unlike typical rewards used in games such as ``wins'' or ``losses'' -- these rewards are invariant to the state's current mover. This means that we can collect rewards from \textit{all} encountered states, rather than only from those corresponding to a specific player. This policy is trained using a discount factor $\gamma = 0.99$.

In self-play, we no longer sample actions proportionally to the visit counts of MCTS, but we sample actions from a mixed distribution with action-probabilities $0.9 \mathcal{M}_s(s, a) + 0.1 \mu(s, a)$. 
A correction for the modified probabilities for a single step requires an IS ratio $\rho(S_t) = \dfrac{\mathcal{M}_{S_t}(S_t, A_t)}{0.9 \mathcal{M}_{S_t}(S_t, A_t) + 0.1 \mu(S_t, A_t)}$. 
As in multi-step off-policy RL settings \cite{Precup2000EligibilityTraces, Precup2001OffPolicyTD}, longer trajectories of multiple time steps with a modified behaviour policy require a product of many such IS ratios. For improved stability -- and to avoid cases where large portions of entire episodes become completely useless when $\mathcal{M}_{S_t}(S_t, A_t) = 0$ but $\mu(S_t, A_t) > 0$ -- we truncate these (products of) IS ratios to always lie in $[0.1, 2]$. This comes at the cost of some bias.

\section{Experiments} \label{Sec:Experiments}

This section describes experiments used to empirically evaluate the effects of weighting states according to episode durations (WED), Prioritized Experience Replay (PER), and Cross-Entropy Exploration (CEE) on the performance of agents with policies trained using ExIt.

\subsection{Setup}

We use fourteen different board games, implemented in the Ludii general game system \cite{Piette2020Ludii}; Amazons, Ard Ri, Breakthrough, English Draughts, Fanorona, Gomoku, Hex, Knightthrough, Konane, Pentalath, Reversi, Surakarta, Tablut, and Yavalath. These are all two-player, deterministic, perfect information board games, but otherwise varied in mechanics and goals. Ard Ri and Tablut are highly asymmetric games.

For each of WED, PER, and CEE, we run a training run of ExIt for 200 games of self-play. We also include a standard ExIt run (without any of the extensions discussed in this paper), an additional run of CEE without performing any IS corrections, and a training run that uses WED, PER, and CEE (without IS) simultaneously. Policies use local patterns \cite{Browne2019StrategicFeatures} as binary features for state-action pairs. We start every training run with a limited set of ``atomic'' features, and add one feature to every feature set after every full game of self-play \cite{Soemers2019BiasingMCTS}. Because we include asymmetric games, we use separate feature sets, separate experience buffers, and train separate feature weights, per player number (or colour). Experience buffers have a maximum capacity of 2500 states. Policies are trained by taking a gradient descent step at the end of every time step in self-play, using a centred variant \cite{Graves2013GeneratingSequences} of RMSProp as optimiser, and batches of 30 states to estimate gradients.

PER uses $\alpha = \beta = 0.5$ for its hyperparameters. These are the default values for PER in the Dopamine framework \cite{Castro2018Dopamine}. In all cases where IS is used for WED, PER, or CEE, we use WIS estimators of the form in \refequation{Eq:WIS} to estimate gradients. The unbiased, higher-variance ordinary IS estimators were found not to perform as well in preliminary experiments.

For every training run, we store checkpoints of feature sets and trained weights after $1$, $51$, $101$, $151$, and $200$ games of self-play, leading to five different versions of each of the following: \textbf{ExIt} (no extensions), \textbf{WED}, \textbf{PER}, \textbf{CEE}, \textbf{CEE (No IS)}, and \textbf{WED + PER + CEE (No IS)}, for a total of 30 trained agents. In evaluation games, we also add two more non-learning agents as benchmarks: \textbf{UCT} (a standard UCT \cite{Browne2012} implementation), and \textbf{MC-GRAVE} (an implementation of GRAVE \cite{Cazenave2015GRAVE} without exploration term in the selection phase), for a total of 32 agents participating in evaluation games.

UCT uses a value of $\sqrt{2}$ for its exploration constant. All of the trained agents use $C_{PUCT} = 2.5$ in \refequation{Eq:PUCT}. All variants of MCTS re-use relevant parts of search trees from previous searches, and run 800 iterations per move -- in training as well as evaluation games. The use of 800 iterations is consistent with AlphaZero \cite{Silver2018AlphaZero}. Value estimates in all variants of MCTS lie in $[-1, 1]$. Unvisited nodes are always estimated to have a value equal to the value estimate of their parent, except in MC-GRAVE where unvisited nodes get a value estimate of $10,000$. In evaluation games, all agents select the action that maximises the visit count (breaking ties randomly).

For every game, we run 120 evaluation matches for every possible (unordered) pair of agents that could be sampled -- with replacement -- from the total pool of 32 agents. Every agent plays each side of its matchup in half of the evaluation games (i.e. 60 out of 120).

\subsection{Results}

\begin{figure}[!t]
    \centering
    \includegraphics{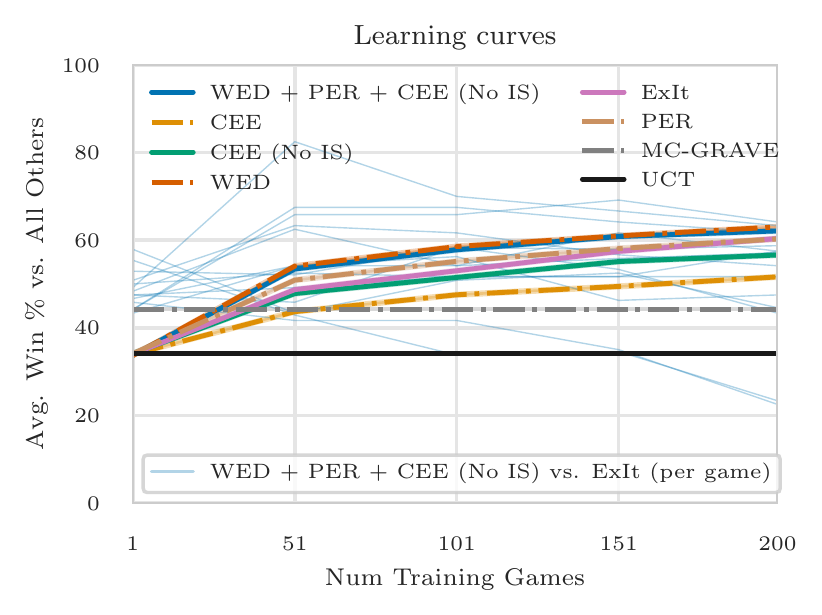}
    \caption{Thick lines depict progression of win percentages, averaged over all fourteen games and all 31 possible opponent agents, including $95\%$ Agresti-Coull confidence intervals for the mean over all those data points. The fourteen thin lines depict win percentages in individual games of \textbf{WED + PER + CEE (No IS)}, against only \textbf{ExIt} with equal numbers of training games.
    }
    \label{Fig:LearningCurves2Player}
\end{figure}

The thick lines in \reffigure{Fig:LearningCurves2Player} depict the average win percentages of each of the 30 different (checkpoints of) learning agents across all games against all $31$ possible opponents. Different checkpoints of the same training run are connected, forming learning curves. The two non-learning agents (UCT and MC-GRAVE) are drawn as horizontal lines. The fourteen thin lines depict similar learning curves for WED + PER + CEE (No IS) for individual games (i.e., not averaged over all games), and only use ExIt at equal training checkpoints as opponent (i.e., not averaged over all opponents).

While these win percentages offer some insight into relative playing strengths, a shortcoming of this metric is that every possible opponent is considered equally important. Suppose that there are three agents $A$, $B$, and $C$. If $A$ outperforms the other two by a small margin, we may consider it to be the strongest agent. But if $B$ (outperformed by $A$) more aggressively exploits the weakest agent $C$, it may be ranked as the top agent by average win percentages. Therefore, we also evaluate our agents using $\alpha$-rank \cite{Omidshafiei2019AlphaRank, LanctotEtAl2019OpenSpiel}. This ranking approach, based on evolutionary game theory, would find that agent $A$ is dominated and would be eliminated from the population of agents in the example above.

We use tables of pairwise win rates as payoff tables for $\alpha$-rank, conducting a sweep over its ranking-intensity hyperparameter $\alpha$ to find sufficiently high values \cite{Omidshafiei2019AlphaRank} for every game. We treat all games as asymmetric games, meaning that $\alpha$-rank does not generate rankings of agents, but rankings of \textit{pairs} of agents corresponding to the two player indices in $2$-player games. In some games the same agent is the top-performing agent for both player numbers, but there are also cases where one agent performs best as Player 1 and another as Player 2.

\begin{table}
\caption{Results for $\alpha$-rank evaluations in fourteen games. The second column shows the number of top ranks that every agent has obtained across the games (with $2$ top ranks available per game, for the $2$ players). The third column shows the strategy masses of the different agents in the stationary distribution of $\alpha$-rank (averaged over games).}
\label{Table:AlphaRank}
\centering
\begin{tabular}{@{}lrr@{}}
\toprule
     Agent & Num. Top Ranks & Avg. Strategy Mass \\
     \midrule
     UCT & $0$ & $0.010$ \\
     MC-GRAVE & $4$ & $0.145$ \\
     ExIt & $3$ & $0.085$ \\
     WED & $9$ & $0.304$ \\
     PER & $2$ & $0.118$ \\
     CEE & $1$ & $0.048$ \\
     CEE (No IS) & $4$ & $0.146$ \\
     WED + PER + CEE (No IS) & $5$ & $0.144$ \\
     \midrule
     \textbf{Total} & $28$ & $1.0$ \\
\bottomrule
\end{tabular}
\end{table}

\reftable{Table:AlphaRank} shows the results of the $\alpha$-rank evaluations. For every agent, we count how often it is present in the top-ranked strategy across all games. There is a total of $28$ top ranks available across fourteen games. For every agent, we also compute the strategy mass of that agent in $\alpha$-rank's stationary distribution over strategies -- averaged over the fourteen games. These two metrics are often correlated, but can still provide different insights. When a single agent clearly outperforms all the others, it achieves the top ranks as well as gaining all the strategy mass in a game. When multiple closely-matched agents outperform each other (e.g., pure strategies in Rock-Paper-Scissors), the strategy mass is more evenly distributed among these agents.

For the trained agents, we add up the top ranks and strategy masses for all the different checkpoints of the same training run. There were only few cases where the final checkpoints were not definitively the strongest agents of their run.

\section{Discussion} \label{Sec:Discussion}

In \reffigure{Fig:LearningCurves2Player}, we see WED and the combination of extensions WED + PER + CEE (No IS) outperforming the ExIt baseline on average, especially for the early checkpoints of $51$ and $101$ training games, but also in later checkpoints to a lesser extent. PER on its own also has a small positive impact in the initial stages of learning. Both variants of CEE are detrimental for performance on average, with the variant that uses IS corrections performing significantly worse than the variant that ignores IS corrections. 

The thin learning curves in \reffigure{Fig:LearningCurves2Player} show that the combination of extensions leads to major improvements in playing strength in the early stages of training in multiple games, with win percentages between $60\%$ and $85\%$ against ExIt with the same amount of training in five out of fourteen games after $51$ training episodes. For other games, the playing strength tends to be closer to even. After $200$ training episodes, there are two games where the regular ExIt has a major advantage in playing strength, but on average the extensions still lead to a minor advantage. For other extensions, we similarly observed that there can be major effects -- both positive and negative -- in individual games, but we omit these plots for visual clarity.

The $\alpha$-rank evaluations in \reftable{Table:AlphaRank} show particularly dominant results for WED, in terms of its number of achieved top ranks as well as average presence in the stationary distributions over agents. This is interesting considering it is also the simplest of all the evaluated extensions of ExIt. PER achieves only two top ranks, but has a high average strategy mass relative to this number of top ranks. This suggests that PER has a relatively stable level of performance; it rarely leads to the best agent, but it is also rarely entirely dominated by other strategies. In contrast, ExIt without any extensions has a relatively low average strategy mass.

\section{Conclusion} \label{Sec:Conclusion}

This paper explores three different extensions for the Expert Iteration (ExIt) self-play training framework, all three of which involve manipulations of the distribution of data that we learn from -- either by modifying the distribution of data that we collect, or by modifying how we sample from it. 

Firstly, we investigated applying importance sampling (IS) corrections based on the durations of episodes in which samples of experience were observed, such that -- in expectation -- we optimise the cross-entropy loss for the distribution of states that we would have collected if we only stored one state for every full game of self-play. We still retain sample efficiency because we do in practice retain \textit{all} states -- IS corrects for this discrepancy between the distribution of collected data, and distribution of data for which we optimise. This is referred to as weighting according to episodes durations (WED).

Secondly, we apply Prioritized Experience Replay (PER) \cite{Schaul2016} to the ExIt training framework. The impact that experienced states may have on our training process is estimated by the differences between expert and apprentice policies for these states, and states that are estimated to be more informative are sampled more frequently. IS ratios are used to correct for bias introduced by this non-uniform sampling.

Thirdly, we use REINFORCE \cite{Williams1992REINFORCE} to train an additional exploratory policy that is rewarded for navigating to states in which there is a large mismatch between expert and apprentice policies. This exploratory policy is mixed with the standard visit-count-based policy when selecting actions during self-play training. This is referred to as Cross-Entropy Exploration (CEE). We evaluate the introduction of this exploration mechanism both with and without applying IS corrections to correct for the modified distribution of experienced states.

An empirical evaluation across fourteen different $2$-player games shows that -- on average -- WED, and a combination of WED + PER + CEE (No IS), lead to policies with stronger performance levels in terms of average win percentage against a pool of $31$ other agents. This difference is primarily noticeable in the early stages of training. This pool of other agents includes earlier and later checkpoints of the same training run, all checkpoints of all other training runs, and two non-training agents (UCT and MC-GRAVE). PER on its own also appears to have a minor advantage in early training stages. Either variant of CEE on its own appears to be detrimental. 

An additional evaluation using the $\alpha$-rank \cite{Omidshafiei2019AlphaRank} method from evolutionary game theory provides additional evidence for some of these conclusions. The $\alpha$-rank evaluation is particularly favourable for WED, but also for other extensions proposed in the paper.

From these results, we conclude that it is worth examining the distributions of experience for which we optimise cross-entropy losses in self-play training processes such as ExIt more closely. Various extensions that maniupulate these distributions show improvements in playing strength when averaged over fourteen games. WED, which is arguably the simplest modification examined in this paper, also appears to have one of the most noticeable impacts on training performance. Effects averaged over all games tend to be small, but we observe major effects in individual games.

For CEE, in this paper we focused on training a policy to explore trajectories that leads to large cross-entropy losses. In future work, it would also be interesting to investigate other forms of targeted exploration \cite{Thrun1992RoleExploration}. For example, a policy that has already been trained in one game may be directly used to diversify the experience collected -- and speed up learning -- in a second game \cite{Madden2004TransferExperience}. Finally, it would be interesting to investigate if there are certain patterns to which extensions provide positive or negative effects in which games.

\section*{Acknowledgment.}

This research was conducted as part of the European Research Council-funded Digital Ludeme Project (ERC Consolidator Grant \#771292), run by Cameron Browne at Maastricht University's Department of Data Science and Knowledge Engineering (DKE). We thank Shayegan Omidshafiei for guidance on the $\alpha$-rank evaluations.

\bibliographystyle{IEEEtran}
\bibliography{IEEEabrv,References}

\end{document}